\RequirePackage[l2tabu, orthodox]{nag}
\documentclass[version=last, pagesize, DIV=calc, twoside=off, bibliography=totoc, fontsize=14pt, a4paper, english]{scrartcl}
\input{glyphtounicode}
\usepackage[T1]{fontenc}
\usepackage[utf8]{inputenc}
\usepackage{newunicodechar}
\usepackage{lmodern}
\usepackage{textcomp}
\DeclareFontShape{OMX}{cmex}{m}{n}{
  <-7.5> cmex7
  <7.5-8.5> cmex8
  <8.5-9.5> cmex9
  <9.5-> cmex10
}{}

\SetSymbolFont{largesymbols}{normal}{OMX}{cmex}{m}{n}
\SetSymbolFont{largesymbols}{bold}  {OMX}{cmex}{m}{n}
\usepackage{booktabs}
\usepackage{calc}
\usepackage{tabularx}

\usepackage{etoolbox} 
\newtoggle{LCpres}
\togglefalse{LCpres}

\usepackage{mathtools} 
	\mathtoolsset{showmanualtags}

\usepackage{listings} 
\usepackage[nolist,smaller,printonlyused]{acronym}
\usepackage{fmtcount}
\usepackage[nodayofweek]{datetime}
\nottoggle{LCpres}{
	\usepackage[textsize=small]{todonotes}
}{
}
\usepackage{authblk}

\iftoggle{LCpres}{
	\usepackage{hyperref}
}{
	\usepackage{hyperref}
}
\hypersetup{breaklinks, colorlinks, linkcolor=black, citecolor=black, urlcolor={blue!80!black}, bookmarksopen}
\iftoggle{LCpres}{
	\hypersetup{hyperfigures}
}{
}

\usepackage{bookmark}

\nottoggle{LCpres}{
	\usepackage{enumitem} 
}{
}
\usepackage{ragged2e} 
\usepackage{siunitx} 
\usepackage{braket} 
\usepackage[sectionbib]{natbib}
\usepackage{doi}

\usepackage{amsmath,amsthm}
\usepackage{amssymb}

\usepackage{environ}
\usepackage[capitalize]{cleveref}
\usepackage{pgf}
\usepackage{pgfplots}
	\usetikzlibrary{babel, matrix, fit, plotmarks, calc, trees, shapes.geometric, positioning, plothandlers, arrows, shapes.multipart}
\pgfplotsset{compat=1.11}
\usepackage{graphicx}

\graphicspath{{graphics/},{graphics-dm/}}
\DeclareGraphicsExtensions{.pdf}
\LetLtxMacro\SavedIncludeGraphics\includegraphics
\AtBeginDocument{
	\def\includegraphics#1#{
		\IncludeGraphicsAux{#1}%
	}%
}
\newcommand*{\IncludeGraphicsAux}[2]{%
	\XeTeXLinkBox{%
		\SavedIncludeGraphics#1{#2}%
	}%
}%

\usepackage{printlen}
\uselengthunit{mm}
\usepackage{scrhack}
\usepackage{mathrsfs}

\iftoggle{LCpres}{
	\usepackage{appendixnumberbeamer}
	\setbeamertemplate{navigation symbols}{} 
	\usepackage{preamble/beamerthemeParisFrance}
	\usefonttheme{professionalfonts}
	\setcounter{tocdepth}{10}

\setbeamertemplate{enumerate item}
{
  \begin{pgfpicture}{-1ex}{-0.65ex}{1ex}{1ex}
    \usebeamercolor[fg]{item projected}
    {\pgftransformscale{1.75}\pgftext{\normalsize\pgfuseshading{bigsphere}}}
    {\pgftransformshift{\pgfpoint{0pt}{0.5pt}}
      \pgftext{\usebeamercolor[fg]{item projected}\usebeamerfont*{item projected}\insertenumlabel}}
  \end{pgfpicture}%
}

\setbeamertemplate{enumerate subitem}
{
  \begin{pgfpicture}{-1ex}{-0.55ex}{1ex}{1ex}
    \usebeamercolor[fg]{subitem projected}
    {\pgftransformscale{1.4}\pgftext{\normalsize\pgfuseshading{bigsphere}}}
    \pgftext{%
      \usebeamercolor[fg]{subitem projected}%
      \usebeamerfont*{subitem projected}%
      \insertsubenumlabel}
  \end{pgfpicture}%
}

\setbeamertemplate{enumerate subsubitem}
{
  \begin{pgfpicture}{-1ex}{-0.55ex}{1ex}{1ex}
    \usebeamercolor[fg]{subsubitem projected}
    {\pgftransformscale{1.4}\pgftext{\normalsize\pgfuseshading{bigsphere}}}
    \pgftext{%
      \usebeamercolor[fg]{subsubitem projected}%
      \usebeamerfont*{subitem projected}%
      \insertsubsubenumlabel}
  \end{pgfpicture}%
}

}{
}

\newcommand{\R}{ℝ}

\newcommand{\powerset}[1]{\mathscr{P}(#1)}
\newcommand{\suchthat}{\;\ifnum\currentgrouptype=16 \middle\fi|\;}

\AtBeginDocument{%
	\renewcommand{\epsilon}{\varepsilon}
}

\newcommand{\allalts}{\mathscr{X}}
\newcommand{\alts}{A}

\newcommand{\prof}{\mathbf{R}}

\DeclareDocumentCommand{\lato}{ O{\prof} O{\alts} }{[#1 \!⟼\! #2]}
\newcommand{\tightoverset}[2]{%
  \mathop{#2}\limits^{\vbox to -.5ex{\kern-0.9ex\hbox{$#1$}\vss}}}
\DeclareDocumentCommand{\latoin}{ O{\prof} O{\alpha} }{[#1 \tightoverset{\in}{⟼} #2]}

\newcommand{\nonemptyor}[2]{\ifthenelse{\equal{#1}{}}{#2}{#1}}

\newcommand{\crits}{\mathcal{G}}

\definecolor{darkgreen}{rgb}{0,0.6,0}
\newcommand{\commentOC}[1]{}

\theoremstyle{remark}
	\newtheorem{examplex}{Example}

\newenvironment{example}{
	\pushQED{\qed}\examplex
}{
	\popQED\endexamplex
}

\crefname{examplex}{Example}{Examples}

\apptocmd{\sloppy}{\hbadness 10000\relax}{}{}

\makeatletter
\patchcmd{\@doi}{dx.doi.org}{doi.org}{}{}
\makeatother

\newlength{\xdescwd}
\makeatletter
\NewEnviron{xdesc}{%
  \setbox0=\vbox{\hbadness=\@M \global\xdescwd=0pt
    \def\item[##1]{%
      \settowidth\@tempdima{\textbf{##1}:}%
      \ifdim\@tempdima>\xdescwd \global\xdescwd=\@tempdima\fi}
  \BODY}
  \begin{description}[leftmargin=\dimexpr\xdescwd+.5em\relax,
    labelindent=0pt,labelsep=.5em,
    labelwidth=\xdescwd,align=left]\BODY\end{description}}
\makeatother

\makeatletter
\newcommand{\boldor}[2]{%
	\ifnum\strcmp{\f@series}{bx}=\z@
		#1%
	\else
		#2%
	\fi
}

\makeatother

\newunicodechar{ℝ}{\mathbb{R}}
\newunicodechar{≤}{\ensuremath{\leq}}
\newunicodechar{≥}{\ensuremath{\geq}}
\newunicodechar{→}{\ifmmode\rightarrow\else\textrightarrow\fi}
\newunicodechar{⇒}{\ensuremath{\Rightarrow}}
\newunicodechar{∪}{\cup}
\newunicodechar{∩}{\cap}
\newunicodechar{¬}{\ifmmode\lnot\else\textlnot\fi}
\newunicodechar{…}{\ifmmode\ldots\else\textellipsis\fi}

\newlength{\GraphsNodeSep}
\newlength{\MCDSCatHeight}
\newlength{\MCDSAltHeight}
\newlength{\MCDSAltSep}
\newlength{\MCDSCatWidth}
\newlength{\MCDSAltWidth}
\newlength{\MCDSEvalRowHeight}
\newlength{\MCDSAltsToCatsSep}
\newlength{\MCDSArrowDownOffset}
\tikzset{/Graphs/dot/.style={
	shape=circle, fill=black, inner sep=0, minimum size=1mm
}}
\tikzset{/MC/D/S/alt/.style={
	shape=rectangle, draw=black, inner sep=0, minimum height=\MCDSAltHeight, minimum width=\MCDSAltWidth
}}
\tikzset{MC/D/S/pref/.style={
	shape=ellipse, draw=gray, thick
}}
\tikzset{/MC/D/S/cat/.style={
	shape=rectangle, draw=black, inner sep=0, minimum height=\MCDSCatHeight, minimum width=\MCDSCatWidth
}}
\tikzset{/MC/D/S/evals matrix/.style={
	matrix, row sep=-\pgflinewidth, column sep=-\pgflinewidth, nodes={shape=rectangle, draw=black, inner sep=0mm, text depth=0.5ex, text height=1em, minimum height=\MCDSEvalRowHeight, minimum width=12mm}, nodes in empty cells, matrix of nodes, inner sep=0mm, outer sep=0mm, row 1/.style={nodes={draw=none, minimum height=0em, text height=, inner ysep=1mm}}
}}

\newlength{\GitCommitSep}
\tikzset{/Git/commit/.style={
	shape=rectangle, draw, minimum width=4em, minimum height=0.6cm
}}
\tikzset{/Git/branch/.style={
	shape=ellipse, draw, red
}}
\tikzset{/Git/head/.style={
	shape=ellipse, draw, fill=yellow
}}

\tikzset{profile matrix/.style={
	matrix of math nodes, column sep=3mm, row sep=2mm, nodes={inner sep=0.5mm, anchor=base}
}}
\tikzset{rank-profile matrix/.style={
	matrix of math nodes, column sep=3mm, row sep=2mm, nodes={anchor=base}, column 1/.style={nodes={inner sep=0.5mm}}, row 1/.style={nodes={inner sep=0.5mm}}
}}
\tikzset{rank-vector/.style={
	draw, rectangle, inner sep=0, outer sep=1mm
}}
\tikzset{isolated rank-vector/.style={
	draw, matrix of math nodes, column sep=3mm, inner sep=0, matrix anchor=base, nodes={anchor=base, inner sep=.33em}, ampersand replacement=\&
}}

\tikzset{/GUI/button/.style={
	rectangle, very thick, rounded corners, draw=black, fill=black!40
}}

\tikzset{/logger/main/.style={
	shape=rectangle, draw=black, inner sep=1ex, minimum height=7mm
}}
\tikzset{/logger/helper/.style={
	shape=rectangle, draw=black, dashed, minimum height=7mm
}}
\tikzset{/logger/helper line/.style={
	<->, draw, dotted
}}

\tikzset{/Beliefs/D/S/attacker/.style={
	shape=rectangle, draw, minimum size=8mm
}}
\tikzset{/Beliefs/D/S/supporter/.style={
	shape=circle, draw
}}

\begin{acronym}
\acro{AMCD}{Aide Multicritère à la Décision}
\acro{ASA}{Argument Strength Assessment}
\acro{DA}{Decision Analysis}
\acro{DM}{Decision Maker}
\acro{DPr}{Deliberated Preferences}
\acro{DRSA}{Dominance-based Rough Set Approach}
\acro{DSS}{Decision Support Systems}
\acrodefplural{DSS}{Decision Support Systems}
\acro{EJOR}{European Journal of Operational Research}
\acro{LNCS}{Lecture Notes in Computer Science}
\acro{MCDA}{Multicriteria Decision Aid}
\acro{MIP}{Mixed Integer Program}
\acro{NCSM}{Non Compensatory Sorting Model}
\acro{PL}{Programme Linéaire}
\acro{PLNE}{Programme Linéaire en Nombres Entiers}
\acro{PM}{Programme Mathématique}
\acro{MP}{Mathematical Program}
\acro{MIP}{Mixed Integer Program}
\acrodefplural{PM}{Programmes Mathématiques}
\acro{PMML}{Predictive Model Markup Language}
\acro{RESS}{Reliability Engineering \& System Safety}
\acro{SMAA}{Stochastic Multicriteria Acceptability Analysis}
\acro{URPDM}{Uncertainty and Robustness in Planning and Decision Making}
\acro{XML}{Extensible Markup Language}
\end{acronym}

\newcommand{\dollars}[1]{\SI{#1}[\$]{}}

\begin{document}
\title{%
	\texorpdfstring{%
		Reasons and Means to Model Preferences as Incomplete%
		\thanks{This work has received support under the program “LABEX MS2T” launched by the French Government and implemented by ANR with the references ANR-11-IDEX-0004-02.
		This is an author version. The final publication is available at Springer via \url{https://doi.org/10.1007/978-3-319-67582-4_2}. The text of this version is identical to the final publication, except for this footnote, a few corrected typos, and a rephrasing of the very last paragraph (page \pageref{epistemic}).
		\textbf{Erratum}: in the first paragraph of \cref{sec:mcdm}, contrary to what has been written, the points of view are not necessarily represented by weak-orders. This is illustrated in \cref{ex:electre} where the relation $C$ may violate transitivity (consider $x_1 = (2, 0, 1)$, $x_2 = (1, 2, 0)$ and $x_3 = (0, 1, 2)$).}
	}{Reasons and Means to Model Preferences as Incomplete%
	}
}
\author{Olivier Cailloux}
\affil{Université Paris-Dauphine, PSL Research University, CNRS, LAMSADE, 75016 PARIS, FRANCE\\
	\href{mailto:olivier.cailloux@dauphine.fr}{olivier.cailloux@dauphine.fr}
}
\author{Sébastien Destercke}
\affil{Sorbonne Université, UMR CNRS 7253 Heudiasyc, \\
	Université de Technologie de Compiègne CS 60319 - 60203 Compiègne cedex, France
}
\hypersetup{
	pdftitle={Reasons and means to model preferences as incomplete},
	pdfsubject={preference modeling},
	pdfkeywords={mcda, risk, decision making, modeling}
}
\date{\formatdate{7}{12}{2017}}
\maketitle

\abstract{Literature involving preferences of artificial agents or human beings often assume their preferences can be represented using a complete transitive binary relation. Much has been written however on different models of preferences. We review some of the reasons that have been put forward to justify more complex modeling, and review some of the techniques that have been proposed to obtain models of such preferences.}


\section{Introduction}\label{sec:intro}
Preferences of agents are usually assumed to be representable with a weak order (a complete and transitive binary relation). We are interested in discussing the completeness assumption. 
	
Preference models are especially important in two fields: choosing an alternative when it is evaluated according to different aspects (multi-criteria decision making, or MCDM), and picking an alternative whose quality depends on states of the world that are uncertainly known (decision making under uncertainty, or DMU).
In MCDM, the common assumption is that the alternatives, i.e., the state of the world, is known without ambiguity, and the difficulty is to determine the structure of the user’s preferences over these well-defined alternatives. In DMU, the alternatives are usually not described over several criteria, but the problem is to recommend an alternative given our uncertainty about the world. 
	
In this paper, we review some reasons to relax preference completeness and modeling approaches (either in MCDM or DMU) that support this relaxation. We discuss in particular reasons to consider that the assumption of completeness is empirically falsified. Although these reasons are not new, we think it is interesting to discuss this question here and now because of (as we perceive it) a relative ignorance of these discussions in research fields that use preference models but are not specialized in preference modeling per se, and because of recent and ongoing advances in analysis of incomplete preferences.
We try to cover a wide scope by discussing some of the goals, assumptions and basic definitions related to preference modeling and reviewing a wide range of techniques for obtaining such models. In counterpart, this review does not claim to be comprehensive and does not provide technical details. To further simplify the discussion, we pretend that the MCDM and DMU contexts are sharply separated. (In reality, it is often possible to cover MCDM contexts while taking uncertainty into account \citep{keeney_decisions_1993}.) We also do not discuss transitivity.

We briefly present the MCDM and DMU settings considering completeness in Section~\ref{sec:review}. \Cref{sec:nordesc} discusses completeness in descriptive and normative approaches (recalling their difference at the same time). Finally, we review models that departs from completeness in Section~\ref{sec:incomp}.
	
\section{Assuming completeness}\label{sec:review}
In this section, we are going to recall the main models that consider completeness and transitivity of preferences as a consequence of natural requirements, if not as pre-requisite of any preference modeling. We will also recall normative views and descriptive views of these concepts. 
	
\subsection{MCDM}
We consider a simple and classical setting in MCDM. We assume that the alternatives are evaluated using a set of criteria $\crits$, each having an evaluation scale $X_g$. The set of all possible alternatives is $\allalts = \prod_{g \in G} X_g$, that is, every combination of evaluations are considered possible. 
We are interested in a preference relation $\succeq$ defined as a binary relation over $\allalts$.

\begin{example}
	Say the \ac{DM} must choose what to plant in her garden. The set of alternatives $\allalts$ are all possible vegetables, the criteria $\crits = \{g_1, g_2, g_3\}$ measure the taste, quantity, and price of each vegetable. 
The scales are $X_{g_1} = \{A, B, C, D\}$, a set of labels, with $x_1$ representing the taste of the vegetable $x \in \allalts$ as considered by the \ac{DM} ($A$ is the worst taste, $D$ the best), $X_{g_2} = [0, 100]$, with $x_2$ representing the number of meals that the \ac{DM} would enjoy if deciding to plant $x$, and $X_{g_3} = \R$, thus $x_3$ indicates the price to pay for planting $x$.
\end{example}
	
Typical approaches in MCDM assume that there is some real-valued function $v: \allalts \to \R$ mapping alternatives to their values, and that $x \succeq y$ iff $v(x) ≥ v(y)$.

\subsection{DMU}
In the simplest form of DMU considered here (SDMR, for Simple Decision Making under Risk), we consider a set $S$ of possible states of the world, a finite set of consequences $C$, and each act $x: S \to C$ is modeled as a function where $x(s)$ is the consequence of performing $x$ when $s$ is the actual state of the world. Define $\allalts$, for simplicity, as all possible or imaginary acts (thus $\allalts = C^S$).
In SDMR, uncertainty is modeled by a probability measure $p$ over the power set of $S$, $\powerset{S}$, thus with $p(s) \in [0, 1]$ indicating the probability of occurence of $s$ (with $s \subseteq S$), and $p(S) = 1$. 
We consider a preference relation $\succeq$ defined as a binary relation over $\allalts$.
Given an act $x$ and a probability measure $p$, it is usually convenient to view $x$ as $p_x$, a probability mass over the consequences: define 
$p_x: C → [0, 1]$ as $p_x(c) = p(x^{-1}(c))$, where $x^{-1}(c)$ designate the set of states in which $x$ leads to the consequence $c$. Such a $p_x$ is usually called a lottery. 

\begin{example}\label{exm:DMU}
	Assume you want to go out and wonder about taking or leaving your umbrella. You consider relevant weather state to be $A=$``shiny” and $B=$“raining”, with $A, B \subseteq S$. We assume $A ∪ B = S$ for simplicity. Two simple actions are $x_1$: “take the umbrella” and $x_2$: “leave the umbrella home”, and the consequences are $c_1$: “encumbered” (when taking your umbrella, irrelevant of the weather), $c_2$: “free” (when leaving your umbrella and weather is $A$), and $c_3$: “wet” (when leaving your umbrella and weather is $B$). Assume the probabilities of the states $A$ and $B$ are $0.2$ and $0.8$. Then the constant act $x_1$ can also be described as $p_{x_1}$ with $p_{x_1}(c_1) = 1$ and $p_{x_1}$ being zero everywhere else, and similarly the act $x_2$ can be associated to $p_{x_2}$ where $p_{x_2}(c_1) = 0$, $p_{x_2}(c_2) = 0.2$, $p_{x_2}(c_3) = 0.8$.
\end{example}
	
In most DMU frameworks, consequences can be mapped to a real-valued reward or utility through a function $u_1: C \to \R$, in which case $u_1(x(s))$ denotes the utility of performing $x$ in state $s$, and acts can be evaluated using a utility function $u: \allalts → \R$ defined as $u(x) = \sum_{s \in S} p(s) u_1(x(s))$, such that $u(x) ≥ u(y)$ iff $x \succeq y$.
It follows from this definition that $u$ and $u_1$ are coherent, in the following sense: given an act $x$ that brings a consequence $c$ with probability one, $u(x) = u_1(c)$.

Expected utility has been justified axiomatically by different authors, the main ones being \citet{savage_foundations_1972}, \citet{definetti_probability_2017} and \citet{von_neumann_theory_2004} (hereafter, vNM). In the \citeauthor{definetti_probability_2017} setting, utilities are given as random variables, and a precise price can be associated to each random variable. That reasoning should be probabilistic and choices made according to expected utility follow from two axioms: linearity and boundedness of those prices. vNM postulate conditions on $\succeq$ ensuring that utility functions $u$ and $u_1$ satisfying the above conditions exist. The axioms assume completeness of the preferences, and the probabilities are assumed to be given. In the Savage setting, both probabilities and expected utility follow from axioms about preferences between acts. In particular, his first axiom (P1) is that any pair of act should be comparable. Completeness is therefore postulated in the axioms, and expected utility and probabilistic reasoning follow from the axioms. 

While these theoretical constructs have set very strong foundations for the use of probabilities, in practice experiments such as the \citet{ellsberg_risk_1961} urn (contradicting Savage sure-thing principle) suggest that people do not always act according to expected utility \citep{maccrimmon_utility_1979}. 
	
Since then, many different extensions have been proposed \citep{wakker_prospect_2010, quiggin_generalized_2012}. Others propose to relax the probabilistic assumption, for instance by considering a possibilistic setting  (e.g., \citet{dubois_qualitative_2003} discuss Savage-like axioms), by considering sets of probabilities such as in decision under ambiguity \citep{gajdos_attitude_2008}, or by simply considering completely missing information, such as \citeauthor{wald_statistical_1992}'s~\citeyearpar{wald_statistical_1992} celebrated maximin criterion. 
	
All models presented thus far assume that $\succeq$ is complete (by which we mean that if $\succeq$ is incomplete, then no suitable function exists in the class of functions admitted by models presented thus far).
	
\section{Questioning completeness}\label{sec:nordesc}
Before discussing the reasonableness of restrictions about $\succeq$, we need to say a word about what those preferences really represent and what the goal of modeling those may be. Indeed, the meaning of completeness depend on whether a descriptive or a normative approach is adopted. In particular, we will later discuss “how much” descriptive one must accept to be in order for the completeness hypothesis to stand.
	
\subsection{Descriptive and normative approaches}
In the descriptive approach to preferences, the goal of the model is to reflect the observed behavior of a \ac{DM}. Typically, a set of sample choices of the \ac{DM} is first collected, say, of choices of food products in his favorite store, and we would then try to obtain the model that best reflects his choice attitude. Or, we would query an individual’s preference about pairs of objects, and then try to build a predictive model on the whole set of possible pairs of alternatives (a method called active learning in the machine learning community). Such a model may be used to predict his behavior, e.g. for marketing or regulation purposes. 

Under the normative approach, the goal is to model the way the \ac{DM} ought to choose rationally. Rationality may corresponds to accepted external norms, or to rules accepted by the \ac{DM} after careful thinking. (In the second case, the term prescriptive or constructive may be used instead of normative, but different authors use these terms differently \citep{roy_decision_1993, tsoukias_concept_2007}; we will stick to the term “normative” as an umbrella.) In both cases, the decision outcome using such approach may differ from empirically observed decisions. Consider as an example a recruiter in an enterprise who wants to model the recruitment procedure. After having collected data, it may appear that for some (possibly unconscious) reason, the recruitment is biased against some particular socio-economic category. The \ac{DM} may then want to find a recruitment strategy that avoids such biases, therefore actively trying to build a model contradicting empirical observations. 
	
\citet{mcclennen_rationality_1990, guala_logic_2000} discuss philosophical grounds for accepting a normative model. \citet{anand_are_1987, mandler_difficult_2001} discuss normative grounds for usual axioms about preferences, including completeness.

Choosing between normative or descriptive approaches is not always easy. For instance recommender systems often adopt a descriptive approach. 
But descriptive approaches will, by design, reflect our cognitive limitations. Those limitations are numerous and sometimes obviously not in agreement with what the \ac{DM} himself would do when thinking more carefully, as will be illustrated in \cref{sec:evidence}. 
Providing (more) normative-based automatic recommendations might help provide sound advices, help increase serendipity, and possibly build trust (or avoid mistrust) in the recommender system. For example, the \ac{DM} might appreciate that the recommender system’s advices protect him from exploitations of the \ac{DM}’s cognitive limitations by merchants. (As an old but known example, “the credit card lobby is said to insist that any price difference between cash and card purchases should be labeled a cash discount rather than a credit surcharge” \citep{tversky_rational_1986}.) 
	
\subsection{Defining and testing incompleteness}
\label{sec:empirical}
Defining and testing incompleteness in preferences requires to define “preference” (and thus $\succeq$), as its everyday usage can be ambiguous: \citet{frankfurt_freedom_1971} gives seven interpretations of “to want to”, and this exercice transposes, \emph{mutatis mutandis}, to the notion of preference.
	
	
Here is what vNM say about the preference relation (we have taken this from the very insightful presentation of the vNM approach by \citet{fishburn_retrospective_1989}):
“It is clear that every measurement – or rather every claim of measurability – must ultimately be based on some immediate sensation, 
which possibly cannot and certainly need not be analyzed any futher.
In the case of utility the immediate sensation of preference – of one object or aggregate of objects as against another – 
provides this basis” (3.1.2);
“Let us for the moment accept the picture of an individual whose system of preferences is all-embracing and complete, i.e. who, for any two objects or rather for any two imagined events, possesses a clear intuition of preference. More precisely
 we expect him, for any two alternative events which are put before him as possibilities, to be able to tell which of the two he prefers.” (3.3.2) (The “events” correspond to our alternatives.)
	
Expanding on vNM, we define that the \ac{DM} \emph{prefers} $a$ to $b$ when expressing an intuitive attraction towards $a$ when presented with $a$ and $b$, or an equal attraction towards $a$ and $b$; and this attraction does not change along a reasonable time span and as well as when irrelevant changes in the context happen. 
Here, we assume that $a, b$ are alternatives in $\allalts$ described by their evaluations on the criteria (in MCDM) or by the relevant probability distributions and consequences (in SDMR), and consider as irrelevant changes anything that does not change those descriptions. 

Under this definition, postulating completeness of $\succeq$ amounts to say that choices of the \ac{DM} will not change along time or when irrelevant changes happen. While this is not the only possible definition (others will be mentioned), it appears reasonable and sufficiently formal to make the condition empirically testable.
	
A first, immediate argument against completeness is that preferences are not stable over even very short period of time, a well-accepted fact in experimental psychology. Quoting \citet{tversky_intransitivity_1969}, individuals “are not perfectly consistent in their choices. When faced with repeated choices between x and y, 
people often choose x in some instances and y in others. Furthermore, such inconsistencies are observed even in the absence of systematic changes in the decision maker’s taste which might be due to learning or sequential effects. It seems, therefore, that the observed 
inconsistencies reflect inherent variability 
or momentary fluctuation 
in the evaluative process.” 
	
This argument may not be strong enough however. In absence of other arguments, one might agree that preferences are in reality incomplete but claim that they may appropriately be \emph{modeled} as complete: a model of complete preferences would simply deviate from time to time from what individuals declare because of (perhaps rare) random fluctuations in their expressions of preferences. 
In order to discuss this hypothesis, we turn to the second (and much more interesting) reason for failure of completeness, which is also brought by the literature in empirical psychology. It appears that preferences change may not be attributed solely to random fluctuations: they change in systematic ways according to changes in the presentation of the alternatives or the context that should have no impact from a normative point of view.
	
\subsection{Empirical evidence of incompleteness}
\label{sec:evidence}
In multicriteria contexts, psychologists have shown systematic differences between the so-called choice and matching elicitation procedures \citep{tversky_contingent_1988}. Assume you want to know which of two alternatives $x, y$ the \ac{DM} prefers, in a problem involving two criteria. You can present both and directly ask for a choice. Alternatively, with the matching procedure, you present $x$ with its two evaluations $g_1(x), g_2(x)$, and $y'$ with only $g_1(y') = g_1(y)$, and ask the \ac{DM} for which value $g_2(y')$ $y'$ would be indifferent to $x$. Assuming $\succeq$ satisfies dominance and transitivity, you then know that $x \succeq y$ iff $g_2(y') ≥ g_2(y)$. Although the two elicitation procedures should be equivalent, the authors confirm the prominence hypothesis stating that the more prominent criterion has more importance in choice than in matching. One of their study confront the subject to a hypothetical choice between two programs for control of a polluted beach. Program $x$ completely cleans up the beach at a yearly cost of \dollars{750 000}; program $y$ partially cleans it up for a yearly cost of \dollars{250 000}. They assume that pollution is the more prominent criterion here, hence expect that $x$ will be chosen more often in choice than in matching. Indeed, 48\% out of the 104 subjects confronted with a choice procedure selected $x$, whereas only 12\% out of the 170 subjects selected it in a matching procedure. Similar effects apply to lotteries in SDMR~\citep{luce_utility_2000}.

This phenomenon is known as preference reversal due to a breach of procedure invariance. Another reversal is the one due to description invariance (or framing effect), showing that preferences can change by changing the descriptions of alternatives. In \citet{tversky_framing_1981}, two groups have to choose a  program to prepare against an epidemic outspring that would result otherwise in 600 deaths. The two groups are presented with the same numeric alternatives $x$ and $y$, but on the first group the alternatives are presented in terms of numbers of life saved, while in the second they are presented in terms of death counts. The experiment shows that preferences differ predictably in the two groups. It is indeed well-known that results are perceived differently depending on their descriptions as losses or gains \citet{thaler_toward_1980}.

Numerous other studies exist that show and discuss preference reversal effects \citetext{\citealp[Ch. 2 (from which we took the two studies described here above)]{deparis_etude_2012}, \citealp{lichtenstein_construction_2006, tversky_causes_1990, kahneman_judgement_1981, kahneman_choices_2000}}. How to best account for and predict preference reversals is still debated, but their existence is consensual \citep{wakker_prospect_2010, birnbaum_empirical_2017}. Some skeptics did try to show that preference reversals could be attributed to deficiencies in the design of the studies, but finally came around \citep{slovic_preference_1983}.
	
This shows that the $\succeq$ relation cannot be expected to be complete given our definition. For some alternatives, individuals may be led to declare different preferences, denoting an absence of a clear, intuitive preference for each pairs of alternatives. When thinking more about the comparison and presented with different views of the same problem, individuals may in some cases change their preference. This has been studied empirically \citep{slovic_who_1974, maccrimmon_utility_1979, lichtenstein_reversals_2006}, and \citet[pp. 101–103]{savage_foundations_1972} famously reported that it happened to him.

One may of course want to preserve completeness of preferences, for example to preserve mathematical and computational simplicity. One way to do so, common in experimental psychology, is to restrict further the frame in which preferences are considered. For instance, \citet{luce_utility_2000} indicates clearly that he studies preferences in terms of choice, not judgment; \citet{maccrimmon_real_1980} exclude some kind of loteries from the scope of the model. In such cases, completeness may well be justified. 
In other settings, such as normative approaches or recommender systems, it is unclear that such reductions should be enforced, as they may be hard to impose in practice or lead to behavior that the user may not desire. 
	
	
We also mention two related interesting articles: 
\citet{deparis_when_2012} study the behavior of individuals when they are allowed to make explicit statements of incomparability; \citet{danan_are_2006} propose to consider that an incomparability is observed whenever the \ac{DM} is ready to pay a small price to postpone the decision.

Another, more evident, reason to be interested in models allowing incompleteness is that it may well be that provided information is insufficient to obtain a fully precise models.
	
The next section describes approaches that allow incomplete preference representations.

\section{Dropping completeness}\label{sec:incomp}
\subsection{Incompleteness in MCDM}
\label{sec:mcdm}
Some approaches in MCDM in the family of outranking methods \citep{roy_multicriteria_1996, greco_multiple_2016, bouyssou_evaluation_2000, bouyssou_evaluation_2006, bouyssou_consolidated_2015} can represent incomparabilities. A much used idea is to take into account two points of view, leading to weak-orders $\succeq^1$ and $\succeq^2$, then define ${\succeq} = {\succeq^1} ∩ {\succeq^2}$. Thus, when the two weak-orders strongly disagree about some pair of objects, the result can declare them incomparable. As an example, consider (a simplification of) the ELECTRE III method (our much simplified description only consider the aspects sufficient to obtain incomparabilities). It builds a concordance relation $C$ that determines whether alternative $x$ is sufficiently better than $y$, by accounting only for the criteria in favor of $x$; and a discordance relation $D$ that determines whether $x$ is so much worst than $y$ on some criterion that $x$ cannot possibly be considered better than $y$ (thus implementing a veto effect). Precise definitions of $C$ and $D$ depend on parameters to be fixed when implementing the method.  Then, the model declares that $x \succeq y$ iff $x C y$ and not $x D y$.

\begin{example}
\label{ex:electre}
Consider $\allalts = \R^3$, each criteria to be maximized, and a model according to which $x C y$ iff $x$ is better than or equal to $y$ for at least two criteria, and $x D y$ iff for some $g$, $y_g - x_g ≥ 2$. Such a model would consider the two alternatives $x = (0, 0, 2)$ and $y = (1, 1, 0)$ as incomparable: neither $x \succeq y$ nor $y \succeq x$ hold.
\end{example}

Such approaches tend to consider incomparabilities as intrinsic to the preferences, since even a completely specified preference could lead to incomparabilities. 

Robust methods in MCDM exist that distinguish conclusions about preferences that hold for sure, given limited preferential information from the \ac{DM}, from conclusions that possibly hold. Such methods typically start from a class $M$ of possible models (similar to hypothesis space in machine learning) assumed to be candidate representative models of the \ac{DM} preferences. A robust method, given a class $M$ and a set of constraints $C$ reducing the set of possibles models (typically preference statements given by the \ac{DM}), will consider that $a$ is necessarily preferred to $b$, $a \succeq^N b$, whenever $a \succeq b$ for all relations $\succeq$ in $M$ that satisfy $C$ \citep{greco_ordinal_2008}.
\begin{example} 
Assume that the only thing you know about the \ac{DM} is that she prefers $x = (0, 0, 2)$ to $y = (0, 4, 0)$, and you assume that $\succeq$ satisfies preferencial independence, meaning that the way two alternatives compare does not change when changing equal values on a given criterion. Thus, $M$ contains all relations that satisfy preferencial independence, and $C$ is the constraint $x \succ y$. You may then conclude that $a = (3, 0, 2)$ is preferred to $b = (3, 4, 0)$, thus, $a \succ^N b$, but you ignore whether $c = (1, 1, 1)$ is preferred to $d = (0, 2, 2)$, thus, $¬(c \succeq^N d)$ and $¬(d \succeq^N c)$.
\end{example} 
In such approach, the relation $\succeq^N$ is able to represent incomparabilites. Incomparabilities stem here from a lack of knowledge, and are not necessarily intrinsic to the modeled preference relation, as in principle one could collect enough constraints $C$ about $M$ to identify a unique compatible relation $\succ$ on a set of alternatives. It is of course also possible to include in $M$ some models that allow for incomparabilities \citep{greco_electregkms:_2011}.

\subsection{Incompleteness in DMU}

As recalled in Section~\ref{sec:intro}, probability theory and expected utility are the most widely used tools when having to decide under uncertainty, and naturally induce completeness of preferences. It should however be noted early scholars were critical about the fact that completeness could hold in practice. 
\citet[p. 630]{von_neumann_theory_1953} for example themselves considered completeness as a strong condition: “it is very dubious, whether the idealization of reality which treats this postulate as a valid one, is appropriate or even convenient”.

Many attempts to relax the completeness axioms does so by considerings axioms leading to deal with sets of utilities and sets of probabilities~\citep{aumann_utility_1962}, entangling together aspects about decision and about information modeling. 
	
\subsubsection{Keeping precise probabilities but not expected utility}

Even when having precise probabilities, there are alternatives to expected utility that induce incomplete preferences. One of them that is particularly interesting is the notion of stochastic dominance~\citep{levy_stochastic_1992}. Assuming that the set of consequences is completely ordered by preference, which we denote by $C = \{c_1, …, c_n\}$ where $c_{i-1}$ is preferred to $c_{i}$, then a lottery $p_x$ is said to stochastically dominate $p_y$ iff, for all $1 ≤ i ≤ n$:
\begin{equation}\label{eq:stodom}
\scalebox{0.8}{
$p_x(\{c_1,\ldots,c_i\})=\sum_{j=1}^i p_x(c_j) ≥ p_y(\{c_1,\ldots,c_i\})=\sum_{j=1}^i p_y(c_j).$
}
\end{equation}
Since \cref{eq:stodom} can be satisfied for some $i$ and not for others, possible incomparabilities immediately follow. 
	
\begin{example}
Consider the set of consequences $C=\{c_1, c_2, c_3\}$ and the following lotteries (induced by different acts $x_1,x_2,x_3$), given in vectorial forms: $p_1=(0.5, 0.3, 0.2)$, $p_2=(0.6, 0.3, 0.1)$ and $p_3=(0.7, 0, 0.3)$. 
Then $x_2$ stochastically dominates $x_1$, while $x_3$ is incomparable to both $x_1$ and $x_2$, according to stochastic dominance.
\end{example}
	
The notion of stochastic dominance has some very attractive properties, as:
\begin{enumerate}
	\item it does not necessitate to define utilities over consequences, and merely requires them to be linearly ordered;
	\item it can be be perceived as a criterion allowing for utilities to be ill-defined, as $p_x$ stochastically dominates $p_y$ if and only if $x$ has a higher expected utility than $y$ for any increasing utility function $u$ defined over $C$. 
\end{enumerate}

\subsubsection{Incompleteness from non-precise probabilities}

In the past few decades, different scholars have challenged the need for precise probabilities associated to classical axiomatics, advocating the use of imprecisely defined prices (expected values) or of imprecisely defined probabilities. To mention but a few:
\begin{itemize} 
	\item \citet{levi_enterprise_1983} advocates the uses of sets of probabilities within a logical interpretation of probabilities;
	\item \citet{walley_statistical_1991} extends the de Finetti axioms by assuming that an agent would give different buying and selling prices for an act, therefore allowing indecision if the price is between these bounds;
	\item \citet{shafer_probability_2005} explores a probabilistic setting centered on the notion of Martingale.
\end{itemize}
Such theories can most of the time be associated to the use of convex sets of probabilities, and give rise to decision rules that extend expected utility but do allow incomparabilities. Once we accept that a convex set $\mathcal{P}$ of probabilities (or a formally equivalent representation) can represent our knowledge, incompleteness may ensue. 

A prototypical way to induce incompleteness between acts from incompleteness in probabilities is to adapt expected utility criterion, and among rules doing so, maximality is a popular one (it is championed by Walley, but is considered as early as the 60's~\citep{aumann_utility_1962}). Given acts $x_1, x_2$, maximality says that 
$$x_1 \succeq x_2 \text{ iff }  u(x_1) ≥ u(x_2) \text{ for all } p \in \mathcal{P}.$$
Maximality reduces to expected utility when $\mathcal{P}$ is a singleton. 

\begin{example}
	Going back to \cref{exm:DMU}, imagine that $x_1$ is indifferent to $x_2$ exactly when $p(A) = p(\text{“shiny”}) = 1/3$. Thus, $u_1(\text{“encumbered”}) = 1/3 u_1(\text{“free”}) + 2/3 u_1(\text{“wet”})$. Then, $x_1$ and $x_2$ will be incomparable according to maximality as soon as $\mathcal{P}$ contains at least one mass where $p(A) < 1/3$, and another where $p(A) > 1/3$.
\end{example}

\commentOC{Il faudrait qu’on puisse exprimer ceci en termes de comparaisons binaires.}
It should be noted that other authors have proposed different rules: for instance \citet{levi_enterprise_1983} recommends to use a decision rule, often called E-admissibility, that does not give rise to an incomplete order between acts, but rather selects all the acts that are Bayes optimal according to at least one probability $p \in \mathcal{P}$. In terms of order, this comes down to consider a set of possible linear ordering, and to retain only those elements that are maximal for at least one of them.


\subsubsection{Working with sets of probabilities and utilities}
Sets of probabilities are helpful to represent incomplete beliefs or lack of information, yet it is natural to also consider cases where the \ac{DM} cannot provide a fully accurate estimation of utilities associated to consequences, or even to completely order them. In some sense, stochastic dominance is an extreme view of such a case, where consequences are ordered but the utility function is left totally unspecified. \commentOC{La découpe que nous proposons ne me semble pas claire à cet endroit. Nous avons déjà parlé ci-dessus d’autres approches qui travaillent avec des ensembles de proba, donc on peut se demander ce qui distingue vraiment cette section-ci. Par exemple pourquoi \og{}maximality\fg{} n’est pas ici ?}
	
	Other works have dealt with partially specified utilities.
	\begin{itemize}
		\item \citet{dubra_expected_2004} represents preferences over lotteries by a set of utility functions. Preference holds whenever the expected utility for the preferred alternative is higher for all utility functions. This idea has been applied in other contexts \citep{ok_utility_2002}.
		\item \citet{dubra_model_2002} propose to view the preference relation as a completion of an intuitive partial preference relation: the \ac{DM} knows intuitively the result of some comparisons, and compute the other ones by applying some reasoning process. They also obtain a preference relation that is representable using a set of utility functions. This approach directly tackles some of the shortcomings described in \cref{sec:empirical}.
		\item \citet{manzini_representation_2008} use a utility function and a vagueness function, representing the preference using intervals of utilities rather than real valued utilities. (Beyond DMU, also using this representation, \citet{masatlioglu_rational_2005} assume that a specific alternative called the status quo alternative is prominently chosen whenever the \ac{DM} faces a choice about which incomparability occur.)
	\end{itemize}

	There exist a few works where both requirements of precise probabilities and utilities are relaxed. This can be traced back at least to \citet{aumann_utility_1962} whose axioms do not require uniqueness of utilities: $x \succ y ⇒ u(x) > u(y)$, without requiring the reverse. More recently, \citet{galaabaatar_subjective_2013} are interested in the Savage-like context where probabilities are unknown and represent an incomplete preference relation in uncertaintly using a set of pairs of probabilities and utilities.

	\section{Incompleteness: absence of knowledge or knowledge of absence?}
	
	We have tried to browse a general picture of reasons why preference modeling should accommodate for incompleteness, and how it can do so in multi-criteria problems and uncertainty modeling. 
	
	One issue that transpired in most of the paper is whether incompleteness should be considered as an intrinsic, or ontic property of the preferences, in which case incomparability expresses a knowledge of absence of relation, or if incompleteness should be considered as an incomplete, epistemic description of a complete order, in which case it expresses an absence of knowledge. This mirrors different views about probability sets (Walley's consider that they model belief, without assuming an existing precise unknown distribution, while robust Bayesians consider the opposite). 
	
\phantomsection
\label{epistemic}
	Our opinion is that both views can be legitimate in different settings, and also that beyond the philosophical interest of distinguishing the two, this can have an important practical impact: knowing that incomparabilities are observable facts may influence strongly our information collection protocol; also, a same piece of information will be interpreted differently. Starting, for simplicity, from an irreflexive and transitive relation $\succ$, and assuming we know somehow that $a$ is a maximal element among three acts $\{a,b,c\}$, in the epistemic interpretation, we would deduce that $a$ is preferred to the other two acts ($a \succ \{b,c\}$), but in the ontic one we could only deduce that $a$ has no element preferred to it ($\neg(b\succ a)$ and $\neg(c\succ a)$).

\bibliography{Survey,seb}

\end{document}